\pdfoutput=1

\documentclass[11pt]{article}

\usepackage[preprint]{acl}

\usepackage{times}
\usepackage{latexsym}
\usepackage{amsmath}
\usepackage{booktabs}
\usepackage{algorithm}
\usepackage{algorithmic}
\usepackage{xcolor}

\usepackage[T1]{fontenc}

\usepackage[utf8]{inputenc}

\usepackage{microtype}

\usepackage{inconsolata}

\usepackage{graphicx}

\usepackage{caption}
\usepackage{subfig}
\title{Threshold Filtering Packing for Supervised Fine-Tuning: Training Related Samples within Packs}

 \author{
    \textbf{Jiancheng Dong\textsuperscript{1}}, 
    \textbf{Lei Jiang\textsuperscript{1}}, 
    \textbf{Wei Jin\textsuperscript{2}}, 
    \textbf{Lu Cheng\textsuperscript{1}} \\
    \textsuperscript{1}University of Illinois Chicago, 
    \textsuperscript{2}Emory University \\
    \texttt{dongjiancheng77@gmail.com, \{ljian43, lucheng\}@uic.edu, wei.jin@emory.edu}
}

\begin{document}
\maketitle
\begin{abstract}
Packing for Supervised Fine-Tuning (SFT) in autoregressive models involves concatenating data points of varying lengths until reaching the designed maximum length to facilitate GPU processing. However, randomly concatenating data points can lead to cross-contamination of sequences due to the significant difference in their subject matter. The mainstream approaches in SFT ensure that each token in the attention calculation phase only focuses on tokens within its own short sequence, without providing additional learning signals for the preceding context. To address these challenges, we introduce Threshold Filtering Packing (TFP), a method that selects samples with related context while maintaining sufficient diversity within the same pack. Our experiments show that TFP offers a simple-to-implement and scalable approach that significantly enhances SFT performance, with observed improvements of up to 7\% on GSM8K, 4\% on HumanEval. Furthermore, results from bias benchmark datasets highlight TFP's promising performance in improving fairness while also boosting prediction accuracy by 15\%.

\end{abstract}

\section{Introduction}

\begin{figure*}[t]
\centering
\includegraphics[width=1\textwidth]{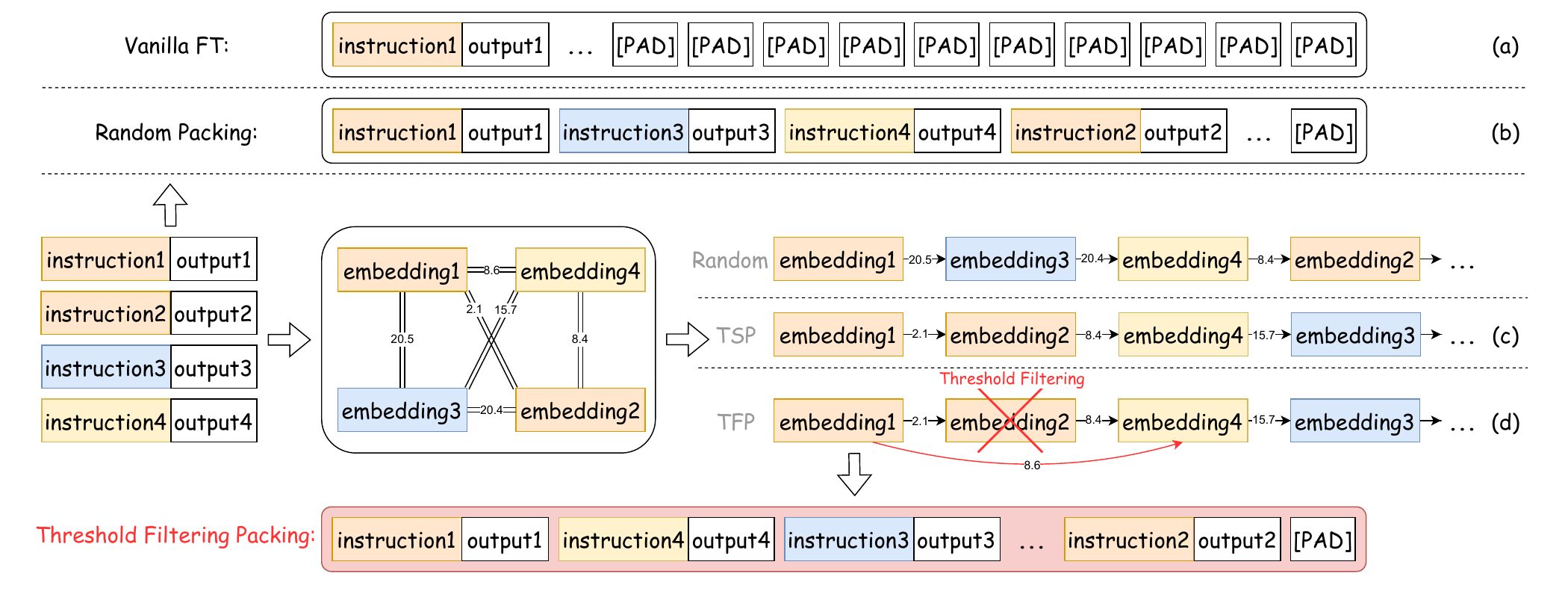} 
\caption{\textbf{Overview of TFP. } 
 Different from vanilla FT (a), which uses “[PAD]” tokens up to the maximum length, and random packing (b), which places randomly shuffled samples in the same pack and may lead to cross-contamination, TFP (d) places related samples in the same pack while applying a threshold on TSP (c) to ensure that these samples are sufficiently distinct. This approach allows models to learn across diverse contexts and prevent overly similar samples from being grouped together.
}
\label{fig1}
\end{figure*}
In Supervised Fine-Tuning (SFT) for large language models (LLMs), sequence lengths can vary substantially, requiring the wrapping of data into tensors to apply matrix operations optimized for CUDA and GPUs \cite{raffel2023exploringlimitstransferlearning}. As illustrated in Figure~\ref{fig1}(a), vanilla fine-tuning pad shorter sequences with special tokens "[PAD]" up to the maximum sequence length. While this ensures uniformity, it introduces inefficiencies by including irrelevant padding tokens in the computation, wasting GPU resources and dilutes the model's learning signal \cite{kundu2024enhancingtrainingefficiencyusing}.

To address this issue, packing sequences has become a common technique in autoregressive transformer models during training and inference to optimize context length and reduce padding \cite{liu2019robertarobustlyoptimizedbert,brown2020languagemodelsfewshotlearners}. This method involves randomly selecting and concatenating data of varying lengths until reaching the designed maximum length. Recent studies suggest that packed data, when batched and processed on multi-GPUs, effectively minimize idle time within each batch \cite{bai2024longalignrecipelongcontext}. 

However, randomly concatenating these data samples (Figure~\ref{fig1}(b)), can result in sequence cross-contamination \cite{krell2022efficientsequencepackingcrosscontamination}. Cross-contamination occurs when predictions for one sequence are influenced by an unrelated sequence, complicating accurate predictions, especially when the subjects differ. For instance, imagine a model tasked with generating a multiplication table, followed by a prompt to list useful expressions in French. If these two sequences are concatenated without proper separation, the model might mix the tasks, producing results like "3 x 2 = Bonjour." This leads to incorrect and confusing outputs, where instructions and contexts become inappropriately blended. Moreover, current SFT pipelines \cite{kundu2024enhancingtrainingefficiencyusing} cause previous samples to provide no signal for predicting the next sample, thereby reducing learning efficiency and negatively impacting the few-shot performance of LLMs.

To address these challenges, in this work, we present Threshold Filtering Packing (TFP), a new packing approach that packs sequences of related yet diverse samples, encouraging context richness and reasoning across sample boundaries. Specifically, we employ a greedy algorithm inspired by the Traveling Salesman Problem (TSP) \cite{article1} to efficiently map out a path for segmentation into multiple packs.
TFP further refine these packs by ensuring that overly similar samples are not grouped together. Setting the threshold is crucial in this context as it allows us to strike a balance between similarity and diversity within each pack, preventing homogeneity and ensuring the robustness of the generated packs. As shown in Figure~\ref{fig1}(c), each sample is first converted to an embedding and then represented as a node in the graph. As shown in Figure~\ref{fig1}(d), TFP employs threshold filtering to ensure each sample is distinct enough from recent ones, preventing the model from merely replicating previous outputs. This method results in packs that provide useful context while avoiding cross-contamination from unrelated texts. 

For experiments, we fine-tune various LLMs on standard instruction fine-tuning datasets and conduct bias-related experiments to evaluate the potential effect of packing on fairness. Additionally, we assess the impact of TFP on computational efficiency for SFT. Our findings indicate that TFP demonstrates superior performance across various models. The bias-related experiments show that TFP offers a flexible operational space, allowing for adjustments in the ratio of sensitive attributes (e.g., race) within packs to effectively manage bias.

\section{Related Work}
\textbf{SFT and Alignment} Fine-tuning is a prevalent strategy to enhance model performance on downstream tasks, evidenced in domains such as coding \cite{wei2023magicoder, luo2024wizardcoder} and arithmetic \cite{yue2024mammoth}.  
Other work has highlighted the importance of consistency in format \cite{liang2024exploringformatconsistencyinstruction}, data quality \cite{chung2022scalinginstructionfinetunedlanguagemodels}, and mixing tasks from different categories \cite{longpre2023flancollectiondesigningdata,iyer2023optimlscalinglanguagemodel} in SFT.
As LLMs evolve, the risk of generating unsafe content increases \cite{su2024api,wang2024decodingtrustcomprehensiveassessmenttrustworthiness}. Established methods for LLM alignment include instruction fine-tuning and reinforcement learning with human feedback (RLHF) \cite{ouyang2022traininglanguagemodelsfollow}. Instruction fine-tuning, also known as SFT, refines pre-trained models using annotated instructional data, often preceding RLHF to aid initial alignment \cite{touvron2023llama2openfoundation}. RLHF employs reinforcement learning to adapt models based on feedback on generated responses. Although RLHF has been pivotal for developing systems like ChatGPT \cite{chatgpt}, isolated instruction fine-tuning can yield comparable outcomes \cite{sun2023principledrivenselfalignmentlanguagemodels} with much less computational and labor costs.\\
\textbf{Packing} While packing is relatively less researched, it is a technique extensively used in frameworks like Hugging Face's SFT Trainer \footnote{\url{https://huggingface.co/docs/trl/sft_trainer}} to expedite inference and training. 
To prevent cross-contamination during self-attention calculation, existing packing approaches involve concatenating sequences into a single tensor and using masking to disregard elements from other sequences during computation \cite{kundu2024enhancingtrainingefficiencyusing}. This method, including variations like LongAlign \cite{bai2024longalignrecipelongcontext} and Prepacking \cite{zhao2024prepackingsimplemethodfast}, enhances training efficiency and minimizes the cross-contamination impact on model performance. However, it necessitates calculating a distinct attention mask for each batch, complicating implementation and increasing memory consumption for masks, which can hinder the effectiveness of flash attention.

In contrast, TFP avoids the need for masking to prevent cross-contamination. Instead, it forms data packs that provide relevant context without additional masking, streamlining implementation and reducing memory overhead.

\section{Threshold Filtering Packing}

The standard practice in packing is to form a pack by concatenating random samples until reaching the maximum context length \cite{zhao2024prepackingsimplemethodfast}. However, randomly concatenated packs do not provide additional learning signals and can lead to cross-contamination of sequences, compared to training on individual samples. In contrast, TFP generates more coherent packs by concatenating related and useful samples together, improving SFT performance and computational efficiency. 

\subsection{Problem Statement}

Given a set of samples $D = \{d_i\}$, where each sample $d_i$ has its instruction converted into an embedding $e_i$, our goal is to organize these samples into packs such that each of them consists of related samples that provide semantic context. Formally, we aim to form a set of packs $P_1 \cdots P_m$ where each pack $P_i = \{d_1, \ldots, d_{k_i}\} \subset D$ and $\bigcup_{i=1}^m P_i = D$.

\subsection{\texorpdfstring{$k$-NN Packing}{$k$-NN Packing}}

An intuitive method for packing is to apply $k$-NN and place each sample along with its retrieved top-$k$ neighbors in the same pack, referred to as $k$-NN packing. This approach maintains sample similarity within each pack but introduces a significant issue: data repetition. Some samples frequently appear as nearest neighbors for multiple others, leading to overlapping packs, i.e., $\exists i \neq j, P_i \cap P_j \neq \emptyset$. For instance, in the CodeAlpaca dataset \cite{codealpaca}, the sample "Construct a loop in Python to display all elements in a list" was included in 94 different packs with $k$-NN Packing, greatly reducing the diversity of pack content.

The data repetition problem can contaminate both individual packs and the entire training process. Within a pack, popular samples that are close to many others in the embedding space do not serve as diverse contexts, increasing the risk of cross-contamination. Across the training process, repeated exposure to these popular samples reduces the diversity of the dataset, potentially leading to overfitting \cite{shi2024incontextpretraininglanguagemodeling}.

\subsection{Threshold Filtering Packing Algorithm}

To address these challenges, we propose packing data samples that provide meaningful context while avoiding repeated selection. Recent research on pretraining language models with related documents has inspired our approach \cite{staniszewski2024structuredpackingllmtraining, shi2024incontextpretraininglanguagemodeling, zhao-etal-2024-analysing}, where similar documents are retrieved to enhance pretraining effectiveness. For Supervised Fine-Tuning datasets, we do not need to consider the degree of the starting node, as the shorter sequence length allows us to easily generate a complete graph. A basic approach involves using a greedy algorithm to select samples with the smallest Euclidean distance to their corresponding embeddings, ensuring that each sample is included only once. This is essentially a greedy algorithm for the TSP \cite{article1}. Intuitively, $k$-NN can select the same data point multiple times when retrieving the nearest neighbors, whereas TSP ensures that each data point is selected only once.

Further, previous studies \cite{yasunaga2023retrievalaugmentedmultimodallanguagemodeling,liu2024makesgooddataalignment} show that maintaining diversity among the input contexts is crucial. We adopt a greedy algorithm for TSP with conditional adjustments and then segment this path into multiple packs to generate packs composed of diverse and relevant samples. TFP is designed to assemble related samples, with Threshold Filtering specifically addressing the challenge of placing overly similar samples in the same pack. As shown in Figure~\ref{fig1}(d), threshold filtering separates overly similar embeddings, such as embedding1 and embedding2, which were initially connected by TSP.

The mathematical formulation of this approach is as follows:

\[
\text{Minimize} \quad \sum_{i=1}^m \sum_{j=1}^{k_i} \sum_{l=j+1}^{k_i} w_{e_{ij}, e_{il}}
\]

\[
\text{subject to} \quad \bigcup_{i=1}^m P_i = D,
\]
\[
P_i \cap P_j = \emptyset \quad \forall i \neq j,
\]
\[
|P_i| = k_i \quad \forall i,
\]
\[
w_{ij} = \sqrt{\sum_{q=1}^n (d_{i,q}- d_{j,q})^2},
\]
\[
w_{e_{ij}, e_{il}} > t \quad \text{for all pairs of recent samples}
\]

Here,
\( m \) represents the number of packs.
\( k_i \) is the number of elements in the \( i \)-th pack.
\( e_{ij} \) refers to the \( j \)-th element in the \( i \)-th pack.
\( w_{ij} \) is the Euclidean distance between elements \( e_i \) and \( e_j \).
\( d_{i,k} \) represents the \( k \)-th feature of element \( e_i \).
\( n \) is the dimensionality of the feature space.
\( t \) is the distance threshold to ensure diversity within a pack.

This objective function minimizes the sum of the Euclidean distances \( w_{e_{ij}, e_{il}} \) between all pairs of elements within each pack \( P_i \). These constraints guarantee that the packs are disjoint, meaning no two packs share any common elements, and ensure that each pack \( P_i \) contains exactly \( k_i \) elements. The Euclidean distance between elements \( e_i \) and \( e_j \) is calculated based on their feature vectors \( d_{i,k} \). Additionally, to ensure diversity within each pack, the constraint \( w_{e_{ij}, e_{il}} > t \) is applied
. This method results in packs that provide useful context while avoiding cross-contamination from unrelated texts.
\begin{algorithm}
\caption{Threshold Filtering Packing}
\begin{algorithmic}[1]
\STATE \textbf{Input}: Instruction embeddings array $E$, distance threshold $t$, the distance threshold number $r$
\STATE \textbf{Output}: A shortest path satisfying the threshold condition
\STATE Initialize $n \gets \text{length of } E$, $visited \gets [\text{False}] \times n$ , $visited[0] \gets \text{True}$
\FOR{$i \in \{1, 2, \ldots, n-1\}$}
    \STATE $next \gets$ SelectNext($E$, $visited$, $t$, $r$)
    \STATE \text{//} SelectNext($E$, $visited$, $t$, $r$) select the nearest unvisited embedding in $E$ while ensuring the distance from the recent $r$ samples is greater than $t$
    \STATE $path \gets path \cup [next]$
    \STATE $visited[next] \gets \text{True}$
\ENDFOR
\RETURN $path$
\end{algorithmic}
\label{a1}
\end{algorithm}


As depicted in Algorithm~\ref{a1}, TFP consists of three primary steps: first, generating sentence embeddings \(E\) for the instruction parts of the samples, then selecting the nearest unvisited embedding in \(E\) while ensuring that the distance from the recent \(r\) samples is greater than distance threshold  \(t\). Subsequently, we use the pack formed by instruction-related samples to train the language model. Since TFP only changes the distribution of data within the pack, it can be seamlessly integrated into existing SFT pipelines for LLMs. As a final step, we traverse the samples along the path and concatenate them to create packs. 

\section{Experiment}
\begin{table*}[ht]
\centering
\resizebox{\textwidth}{!}{
\begin{tabular}{lcccccccccccc}
\toprule
 & \multicolumn{3}{c}{\textbf{Llama2-7B}} & \multicolumn{3}{c}{\textbf{Llama3-8B}} & \multicolumn{3}{c}{\textbf{Mistral-7B}} \\
\cmidrule(lr){2-4} \cmidrule(lr){5-7} \cmidrule(lr){8-10}
\textbf{Method} & \textbf{WR} & \textbf{HumanEval} & \textbf{GSM8K} & \textbf{WR} & \textbf{HumanEval} & \textbf{GSM8K} & \textbf{WR} & \textbf{HumanEval} & \textbf{GSM8K} \\
\midrule
Vanilla FT & 48.2 \footnotesize{$\pm$ 0.4} & 19.5 \footnotesize{$\pm$ 0.3} & 26.2 \footnotesize{$\pm$ 0.0} & 51.2 \footnotesize{$\pm$ 0.5} & 38.4 \footnotesize{$\pm$ 0.0} & 61.8 \footnotesize{$\pm$ 0.2} & 62.9 \footnotesize{$\pm$ 0.6} & 35.4 \footnotesize{$\pm$ 0.3} & 59.7 \footnotesize{$\pm$ 0.5} \\
Sorted batching & 48.2 \footnotesize{$\pm$ 0.5} & 20.1 \footnotesize{$\pm$ 0.3} & 26.5 \footnotesize{$\pm$ 0.0} & 51.8 \footnotesize{$\pm$ 0.6} & 37.2 \footnotesize{$\pm$ 0.4} & 62.0 \footnotesize{$\pm$ 0.6} & 61.2 \footnotesize{$\pm$ 0.3} & 34.1 \footnotesize{$\pm$ 0.5} & 61.0 \footnotesize{$\pm$ 0.1} \\
Random packing & 47.1 \footnotesize{$\pm$ 0.3} & 19.5 \footnotesize{$\pm$ 0.3} & 26.1 \footnotesize{$\pm$ 0.4} & 51.7 \footnotesize{$\pm$ 0.5} & 37.8 \footnotesize{$\pm$ 0.3} & 62.1 \footnotesize{$\pm$ 0.6} & 62.4 \footnotesize{$\pm$ 0.0} & 34.8 \footnotesize{$\pm$ 0.6} & 59.1 \footnotesize{$\pm$ 0.4} \\
Random packing (mask) & 47.6 \footnotesize{$\pm$ 0.5} & 19.5 \footnotesize{$\pm$ 0.6} & 26.1 \footnotesize{$\pm$ 0.2} & 52.4 \footnotesize{$\pm$ 0.0} & 37.8 \footnotesize{$\pm$ 0.3} & 62.5 \footnotesize{$\pm$ 0.4} & \textbf{63.5 \footnotesize{$\pm$ 0.2}} & 35.4 \footnotesize{$\pm$ 0.5} & 59.2 \footnotesize{$\pm$ 0.1} \\
Packing+loss weighting & 47.1 \footnotesize{$\pm$ 0.6} & 18.9 \footnotesize{$\pm$ 0.4} & 25.8 \footnotesize{$\pm$ 0.0} & 51.2 \footnotesize{$\pm$ 0.3} & 38.4 \footnotesize{$\pm$ 0.0} & 60.9 \footnotesize{$\pm$ 0.3} & 60.6 \footnotesize{$\pm$ 0.4} & 34.8 \footnotesize{$\pm$ 0.5} & 59.5 \footnotesize{$\pm$ 0.0} \\
$k$-NN packing & 45.3 \footnotesize{$\pm$ 0.0} & 15.9 \footnotesize{$\pm$ 0.6} & 29.3 \footnotesize{$\pm$ 0.3} & 48.8 \footnotesize{$\pm$ 0.4} & 36.0 \footnotesize{$\pm$ 0.0} & 59.5 \footnotesize{$\pm$ 0.5} & 55.3 \footnotesize{$\pm$ 0.4} & 34.1 \footnotesize{$\pm$ 0.3} & 57.2 \footnotesize{$\pm$ 0.2} \\
TFP & \textbf{51.2 \footnotesize{$\pm$ 0.2}} & \textbf{22.6 \footnotesize{$\pm$ 0.3}} & \textbf{33.6 \footnotesize{$\pm$ 0.4}} & \textbf{54.1 \footnotesize{$\pm$ 0.6}} & \textbf{42.7 \footnotesize{$\pm$ 0.5}} & \textbf{66.7 \footnotesize{$\pm$ 0.3}} & \textbf{63.5 \footnotesize{$\pm$ 0.0}} & \textbf{38.4 \footnotesize{$\pm$ 0.5}} & \textbf{64.1 \footnotesize{$\pm$ 0.4}} \\
\bottomrule
\end{tabular}

}
\caption{\textbf{Comparison of different methods and training datasets: Alpaca, CodeAlpaca, and GSM8K, with results represented by WR, HumanEval, and GSM8K, respectively.} In the table, we follow the most common few-shot settings: using Win rate judged by PandaLM and a 0-shot setting for HumanEval, while the GSM8K task uses a 4-shot setting.}
\label{comparison}
\end{table*}

We evaluate the proposed approach from three aspects: (1) \textit{Performance Comparison}: We compare TFP with various LLMs fine-tuned on common instruction fine-tuning datasets under both zero-shot and few-shot settings. (2) \textit{Bias and Fairness}: Inspired by previous research \cite{wang2024decodingtrustcomprehensiveassessmenttrustworthiness} that studies the impact of the ratio of different demographic groups in in-context learning (ICL) on LLM fairness, we investigate how adjusting the ratio in each pack during SFT can influence the bias and fairness of LLMs. (3) \textit{Efficiency}: We study how various SFT methods influence computational efficiency on different GPU setups.

\subsection{Experimental Setup} \label{Setup}
\textbf{Datasets.} We use commonly adopted datasets for instruction fine-tuning, which include tasks related to helpfulness, code-generation capabilities, and mathematical reasoning: (1) {Alpaca dataset}, generated from the Self-Instruct method \cite{wang2023selfinstructaligninglanguagemodels} via the text-davinci-003 model 
\cite{Buruk_2023}, covering various tasks such as arithmetic, coding, and question-answering. (2) CodeAlpaca dataset\cite{codealpaca}, which aims to build and share an instruction-following LLaMA model for code generation. (3) GSM8K dataset \cite{cobbe2021gsm8k}, curated to examine mathematical reasoning capabilities, comprises 8.8k high-quality arithmetic word problems designed at the grade school level. 

For the fairness-related experiments, we use the Jigsaw Unintended Bias in Toxicity Classification task \cite{jigsaw} and Adult dataset~\cite{misc_adult_2}. The Jigsaw Unintended Bias in Toxicity Classification task involves performing toxicity classification on comment texts published by the Civil Comments platform. It contains human-annotated demographic information such as race, gender, and religion. The goal is to ensure that models make predictions based on the text toxicity, rather than demographic information included in the text.  We use race (Black and non-Black) as the protected attributes. The Adult dataset is a tabular dataset that includes 14 attributes of a person (e.g., age and education level) as input, to predict whether the person's income exceeds \$50k per year. We evaluate the fairness of fine-tuned models based on the sensitive attribute of sex, specifically comparing “male” and “female”.\\
\textbf{Baselines.} We compare TFP with the following baselines:

(1) Vanilla fine-tuning: This method appends special padding tokens to shorter prompts to match the maximum length within a batch. Huggingface’s inference framework \cite{wolf2020huggingfacestransformersstateoftheartnatural} generates corresponding attention masks to ensure the language model disregards the padded tokens during computation to handle prompts of variable lengths.

(2) Sorted batching \cite{bai2024longalignrecipelongcontext}: This approach
sorts inputs by length and samples batches to minimize padding.  As a result, each batch consists entirely of either long or short sequences.

(3) Random packing: Random packing involves concatenating data of varying lengths randomly until reaching the maximum length \cite{brown2020languagemodelsfewshotlearners}. 

(4) Random packing (mask): A variant of random packing that uses masking to prevent cross-contamination between different sequences within the same pack during self-attention calculations. 

(5) Packing + loss weighting \cite{bai2024longalignrecipelongcontext}: A typical packing strategy skews towards longer sequences and those with more target tokens, as packs with fewer sequences or more target tokens disproportionately influence the final loss, especially for datasets designed for long contexts. This method ensures equal loss weighting for each sequence.

(6) $k$-NN packing: In this method, each sample is directly placed together with its retrieved top-k samples in the same pack. \\
\textbf{Evaluation Metrics.} We follow commonly used protocols \cite{luo2024wizardcoder, yue2024mammoth, ge2024clusteringrankingdiversitypreservedinstruction} to evaluate SFT in LLMs. Specifically, we use PandaLM~\cite{PandaLM,pandalm2024} to evaluate the helpfulness of various models. PandaLM provides reproducible and automated comparisons between different LLMs. By providing PandaLM with the same context, it can compare the responses of different LLMs, offer reasons for the decisions, and provide a reference answer. We report the win rate (WR), which is the proportion of instances where the responses are favored over those produced by GPT-3.5 \cite{brown2020languagemodelsfewshotlearners}. The code generation skills are enhanced using the CodeAlpaca dataset \cite{codealpaca}, while evaluation is conducted using the HumanEval dataset \cite{chen2021codex}. GSM8K dataset \cite{cobbe2021gsm8k} uses its own test set. We followed the most common few-shot settings: using Win rate judged by PandaLM and a 0-shot setting for HumanEval, while the GSM8K task uses a 4-shot setting.

We utilize the Llama2-7B \cite{touvron2023llama2openfoundation}, Llama3-8B \cite{llama3modelcard}, and Mistral-7B \cite{jiang2023mistral7b} as the base LLM in our experiments. Due to limited computation resources, we employ the QLoRA technique \cite{dettmers2023qloraefficientfinetuningquantized} in all fine-tuning experiments. To ensure fair comparison, we maintain consistency in nearly all hyperparameters across all methods. For all results below, we run the experiments five times and report the mean and standard deviations for all compared methods.

\subsection{Results}

\begin{table*}[ht]
\centering
\resizebox{\textwidth}{!}{
\begin{tabular}{lcccccccccc}
\toprule
\textbf{Method} & \multicolumn{3}{c}{\textbf{0-shot}} & \multicolumn{3}{c}{\textbf{4-shot}}& \multicolumn{3}{c}{\textbf{32-shot}} \\
\cmidrule(lr){2-4} \cmidrule(lr){5-7} \cmidrule(lr){8-10} 
 & \textbf{ACC} $\uparrow$& \textbf{$M_{dpd}$} $\downarrow$ & \textbf{$M_{eod}$} $\downarrow$ & \textbf{ACC} $\uparrow$ & \textbf{$M_{dpd}$}$\downarrow$ & \textbf{$M_{eod}$}$\downarrow$ & \textbf{ACC} $\uparrow$& \textbf{$M_{dpd}$} $\downarrow$& \textbf{$M_{eod}$}$\downarrow$  \\
\midrule

Vanilla FT & 0.84 \footnotesize{$\pm$ 0.01} & 0.10 \footnotesize{$\pm$ 0.00} & 0.12 \footnotesize{$\pm$ 0.01} & 0.75 \footnotesize{$\pm$ 0.02} & 0.08 \footnotesize{$\pm$ 0.01} & 0.10 \footnotesize{$\pm$ 0.02} & 0.73 \footnotesize{$\pm$ 0.01} & 0.08 \footnotesize{$\pm$ 0.00} & 0.16 \footnotesize{$\pm$ 0.02} \\
Random packing & 0.78 \footnotesize{$\pm$ 0.02} & 0.10 \footnotesize{$\pm$ 0.01} & 0.14 \footnotesize{$\pm$ 0.02} & 0.73 \footnotesize{$\pm$ 0.02} & 0.10 \footnotesize{$\pm$ 0.01} & 0.14 \footnotesize{$\pm$ 0.02} &  0.74 \footnotesize{$\pm$ 0.02} & 0.10 \footnotesize{$\pm$ 0.01} & 0.14 \footnotesize{$\pm$ 0.02} \\
Random packing (mask) & 0.84 \footnotesize{$\pm$ 0.00} & 0.08 \footnotesize{$\pm$ 0.01} & 0.12 \footnotesize{$\pm$ 0.02} & 0.71 \footnotesize{$\pm$ 0.02} & 0.08 \footnotesize{$\pm$ 0.01} & 0.12 \footnotesize{$\pm$ 0.02} &  0.72 \footnotesize{$\pm$ 0.01} & 0.08 \footnotesize{$\pm$ 0.01} & 0.12 \footnotesize{$\pm$ 0.02} \\
Balanced ratio & 0.71 \footnotesize{$\pm$ 0.02} & \textbf{0.04 \footnotesize{$\pm$ 0.01}} & \textbf{0.08 \footnotesize{$\pm$ 0.01}} & 0.57 \footnotesize{$\pm$ 0.02} & 0.06 \footnotesize{$\pm$ 0.01} & 0.12 \footnotesize{$\pm$ 0.00} &  0.64 \footnotesize{$\pm$ 0.02} & 0.03 \footnotesize{$\pm$ 0.01} & 0.06 \footnotesize{$\pm$ 0.01} \\
Resampling & 0.85 \footnotesize{$\pm$ 0.01} & 0.11 \footnotesize{$\pm$ 0.02} & 0.16 \footnotesize{$\pm$ 0.00} & 0.66 \footnotesize{$\pm$ 0.02} & 0.07 \footnotesize{$\pm$ 0.01} & 0.14 \footnotesize{$\pm$ 0.02} &  0.71 \footnotesize{$\pm$ 0.02} & 0.14 \footnotesize{$\pm$ 0.01} & 0.28 \footnotesize{$\pm$ 0.02} \\
TFP & \textbf{0.87 \footnotesize{$\pm$ 0.01}} & 0.06 \footnotesize{$\pm$ 0.01} & 0.10 \footnotesize{$\pm$ 0.02} & 0.77 \footnotesize{$\pm$ 0.02} & 0.10 \footnotesize{$\pm$ 0.01} & 0.24 \footnotesize{$\pm$ 0.02} & 0.81 \footnotesize{$\pm$ 0.01} & \textbf{0.01 \footnotesize{$\pm$ 0.01}} & \textbf{0.02 \footnotesize{$\pm$ 0.00}} \\
TFP (Balanced) & 0.85 \footnotesize{$\pm$ 0.01} & \textbf{0.04 \footnotesize{$\pm$ 0.01}} & 0.10 \footnotesize{$\pm$ 0.01} & \textbf{0.78 \footnotesize{$\pm$ 0.01}} & \textbf{0.01 \footnotesize{$\pm$ 0.00}} & \textbf{0.04 \footnotesize{$\pm$ 0.01}} & 0.80 \footnotesize{$\pm$ 0.01} & \textbf{0.01 \footnotesize{$\pm$ 0.01}} & \textbf{0.02 \footnotesize{$\pm$ 0.01}} \\
TFP (Resampling) & \textbf{0.87 \footnotesize{$\pm$ 0.01}} & 0.08 \footnotesize{$\pm$ 0.01} & 0.12 \footnotesize{$\pm$ 0.01} & \textbf{0.78 \footnotesize{$\pm$ 0.01}} & 0.06 \footnotesize{$\pm$ 0.01} & 0.12 \footnotesize{$\pm$ 0.01} & \textbf{0.83 \footnotesize{$\pm$ 0.01}} & 0.05 \footnotesize{$\pm$ 0.01} & 0.06 \footnotesize{$\pm$ 0.00} \\
\bottomrule
\end{tabular}

}
\caption{Accuracy and group fairness metrics on Llama3-8B for the Jigsaw Dataset with balanced few-shots.}
\label{Jigsaw}
\end{table*}

\begin{table*}[ht]
\centering
\resizebox{\textwidth}{!}{
\begin{tabular}{lcccccccccc}
\toprule
\textbf{Method} & \multicolumn{3}{c}{\textbf{0-shot}} & \multicolumn{3}{c}{\textbf{4-shot}}& \multicolumn{3}{c}{\textbf{32-shot}} \\
\cmidrule(lr){2-4} \cmidrule(lr){5-7} \cmidrule(lr){8-10} 
 & \textbf{ACC} $\uparrow$& \textbf{$M_{dpd}$} $\downarrow$& \textbf{$M_{eod}$} $\downarrow$& \textbf{ACC} $\uparrow$& \textbf{$M_{dpd}$} $\downarrow$& \textbf{$M_{eod}$}$\downarrow$ & \textbf{ACC} $\uparrow$& \textbf{$M_{dpd}$}$\downarrow$ & \textbf{$M_{eod}$} $\downarrow$ \\
\midrule

Vanilla FT & 0.78 \footnotesize{$\pm$ 0.02} & 0.26 \footnotesize{$\pm$ 0.03} & 0.44 \footnotesize{$\pm$ 0.04} & 0.67 \footnotesize{$\pm$ 0.02} & 0.06 \footnotesize{$\pm$ 0.01} & 0.09 \footnotesize{$\pm$ 0.02} & 0.54 \footnotesize{$\pm$ 0.03} & 0.04 \footnotesize{$\pm$ 0.01} & 0.08 \footnotesize{$\pm$ 0.02} \\
Random packing & 0.76 \footnotesize{$\pm$ 0.03} & 0.25 \footnotesize{$\pm$ 0.02} & 0.42 \footnotesize{$\pm$ 0.04} & 0.50 \footnotesize{$\pm$ 0.00} & 0.00 \footnotesize{$\pm$ 0.00} & 0.00 \footnotesize{$\pm$ 0.00} & 0.50 \footnotesize{$\pm$ 0.00} & 0.00 \footnotesize{$\pm$ 0.00} & 0.00 \footnotesize{$\pm$ 0.00} \\
Random packing (mask) & 0.78 \footnotesize{$\pm$ 0.02} & 0.25 \footnotesize{$\pm$ 0.02} & 0.44 \footnotesize{$\pm$ 0.04} & 0.65 \footnotesize{$\pm$ 0.02} & 0.06 \footnotesize{$\pm$ 0.01} & 0.08 \footnotesize{$\pm$ 0.02} & 0.56 \footnotesize{$\pm$ 0.02} & 0.06 \footnotesize{$\pm$ 0.01} & 0.08 \footnotesize{$\pm$ 0.02} \\
Balanced ratio & 0.78 \footnotesize{$\pm$ 0.02} & 0.28 \footnotesize{$\pm$ 0.03} & 0.40 \footnotesize{$\pm$ 0.03} & 0.64 \footnotesize{$\pm$ 0.02} & 0.03 \footnotesize{$\pm$ 0.01} & 0.06 \footnotesize{$\pm$ 0.02} & 0.56 \footnotesize{$\pm$ 0.02} & 0.02 \footnotesize{$\pm$ 0.01} & 0.02 \footnotesize{$\pm$ 0.01} \\
Resampling & 0.72 \footnotesize{$\pm$ 0.03} & 0.21 \footnotesize{$\pm$ 0.02} & 0.36 \footnotesize{$\pm$ 0.03} & 0.58 \footnotesize{$\pm$ 0.03} & 0.04 \footnotesize{$\pm$ 0.01} & 0.06 \footnotesize{$\pm$ 0.02} & 0.50 \footnotesize{$\pm$ 0.00} & 0.00 \footnotesize{$\pm$ 0.00} & 0.00 \footnotesize{$\pm$ 0.00} \\
TFP & 0.80 \footnotesize{$\pm$ 0.02} & 0.22 \footnotesize{$\pm$ 0.02} & 0.32 \footnotesize{$\pm$ 0.03} & \textbf{0.81 \footnotesize{$\pm$ 0.02}} & 0.08 \footnotesize{$\pm$ 0.01} & 0.10 \footnotesize{$\pm$ 0.02} & 0.78 \footnotesize{$\pm$ 0.02} & 0.02 \footnotesize{$\pm$ 0.01} & 0.04 \footnotesize{$\pm$ 0.02} \\
TFP (balanced) & 0.80 \footnotesize{$\pm$ 0.02} & 0.11 \footnotesize{$\pm$ 0.02} & 0.16 \footnotesize{$\pm$ 0.03} & 0.78 \footnotesize{$\pm$ 0.02} & \textbf{0.02 \footnotesize{$\pm$ 0.01}} & \textbf{0.02 \footnotesize{$\pm$ 0.01}} & 0.74 \footnotesize{$\pm$ 0.02} & \textbf{0.01 \footnotesize{$\pm$ 0.01}} & \textbf{0.02 \footnotesize{$\pm$ 0.01}} \\
TFP (resampling) & \textbf{0.81 \footnotesize{$\pm$ 0.02}} & \textbf{0.10 \footnotesize{$\pm$ 0.01}} & \textbf{0.16 \footnotesize{$\pm$ 0.02}} & 0.80 \footnotesize{$\pm$ 0.02} & 0.14 \footnotesize{$\pm$ 0.02} & 0.18 \footnotesize{$\pm$ 0.03} & \textbf{0.79 \footnotesize{$\pm$ 0.02}} & 0.15 \footnotesize{$\pm$ 0.02} & 0.24 \footnotesize{$\pm$ 0.03} \\
\bottomrule
\end{tabular}

}
\caption{Accuracy and group fairness metrics on Llama3-8B for the Adult Dataset with balanced few-shots.}
\label{Llama3}
\end{table*}
\subsubsection*{Comparisons of Various Packing Methods in Instruction Fine-tuning}

Table~\ref{comparison} displays the results of fine-tuning on three downstream datasets:  Alpaca \cite{alpaca}, CodeAlpaca\cite{codealpaca}, and GSM8K \cite{cobbe2021gsm8k}. We have the following key observations: 

(1)
TFP consistently outperforms the best of other baselines across various base LLMs, achieving improvements of up to 7\% on GSM8K, 4\% on HumanEval, and 3\% on Alpaca. This suggests that TFP learns more effectively about answering questions 
from the related data within the pack.
However, naive packing strategies (random packing, random packing (mask)) fail to improve performance and can even degrade it.

(2) Comparing TFP with $k$-NN packing, we find that $k$-NN packing can lead to performance declines, especially in well-trained models like Llama3-8B and Mistral-7B, whereas TFP consistently outperforms. Although Llama2-7B showed some improvement on GSM8K with $k$-NN packing, as inferred from previous work \cite{yuan2023scalingrelationshiplearningmathematical}, these gains are likely due to repeated training effects rather than genuine generalization. In contrast, TFP enhances diversity and relevance within packs, leading to better performance without overfitting, underscoring its robustness. 

(3) Sorted batching and loss weighting methods, which are designed for long context, prove less effective on shorter datasets like GSM8K, where the average sequence length of around 500 tokens is far below the model's maximum length. As highlighted in previous work \cite{bai2024longalignrecipelongcontext}, sorted batching can introduce bias in data distribution across batches, where entire batches consist of either long or short sequences, potentially disrupting the optimization process during stochastic gradient descent.


\subsubsection*{Impact of TFP on Fairness}

\begin{table*}[ht]
\centering
\resizebox{\textwidth}{!}{
\begin{tabular}{@{}lccccccccc@{}}
\toprule
 & \multicolumn{3}{c}{\textbf{2*L40S}} & \multicolumn{3}{c}{\textbf{1*3090}} \\
\cmidrule(lr){2-4} \cmidrule(lr){5-7}
\textbf{Method} & \textbf{LLaMA2 7B} & \textbf{LLaMA3 8B} & \textbf{Mistral 7B} & \textbf{LLaMA2 7B} & \textbf{LLaMA3 8B} & \textbf{Mistral 7B} \\
\midrule
Vanilla FT & 1.73 & 1.05 & 1.15 & 4.68 & 4.43 & 5.25 \\
Sorted batching & 1.70 & 1.05 & 1.08 & 4.68 & 4.42 & 5.22 \\
Random packing & \textbf{0.37} & 0.40 & 0.40 & 2.53 & 2.03 & 2.58 \\
Random packing (mask) & 0.40 & 0.38 & 0.40 & 2.57 & 2.50 & 2.62 \\
Packing + loss weighting & 0.42 & 0.43 & 0.45 & \textbf{2.48} & 2.07 & \textbf{2.55} \\
$k$-NN packing & 1.57 & 1.35 & 1.42 & 9.68 & 9.40 & 9.87 \\
TFP & 0.40 & \textbf{0.37} & \textbf{0.38} & 2.57 & \textbf{2.02} & \textbf{2.55} \\
\bottomrule
\end{tabular}}
\caption{Training time (hrs) on 2*L40S and 1*3090 under different training methods.}
\label{tab:llama3-8b_methods_combined}
\end{table*}

Our method, which groups similar data together into packs, can sometimes lead to imbalanced representation across demographic groups, potentially amplifying biases. To address this, we explore two approaches: TFP (balanced) and TFP (resampling), both designed to achieve balance while maintaining text relevance. TFP (balanced) ensures that data with different sensitive attributes are evenly distributed within each pack, continuing this process until samples from one demographic group are exhausted. In contrast, TFP (resampling) addresses imbalances by resampling underrepresented data, ensuring all packs have a balanced representation of attributes, even when certain categories have fewer samples. For comparison, we apply balanced ratio and resampling methods to the data's default sequence as controls, referred to as \textit{Balancing ratio} and \textit{Resampling}, respectively.

Inspired by DecodingTrust \cite{wang2024decodingtrustcomprehensiveassessmenttrustworthiness}, which demonstrates that a balanced ratio across groups in ICL can improve LLM fairness, we examine whether maintaining a balanced ratio of demographic groups within packs during SFT affects LLM fairness in classification tasks. We report prediction accuracy (ACC) and two common fairness metrics used in classification: equalized odds difference (eod) \cite{hardt2016equalityopportunitysupervisedlearning} and demographic parity difference (dpd) \cite{pmlr-v28-zemel13}. Detailed definitions of these metrics are provided in Appendix \ref{fair}. We use 0-shot and balanced 32-shot ICL settings  for evaluation following the experimental design of DecodingTrust, and we also explore the effects of a small number of samples with a balanced 4-shot setting. Experiments in Tables \ref{Jigsaw} and \ref{Llama3} were conducted on text and tabular datasets respectively, from which we have the following three observations. 

(1)
TFP (balanced) excels in fairness tasks for both text and tabular data, especially when the number of balanced ICL few shot examples is large. We believe this is due the compounded effect of both balanced group ratio within the pack and the ICL demonstrations. The original TFP does not excel in fairness in 0-shot settings due to imbalanced data across social groups. TFP (resampling) results in instability due to the repeated sampling of same samples. By providing the model with hard negative examples (i.e. closely positioned samples with differing labels) within a pack, TFP enables more efficient learning from data with similar texts but different labels. Balancing the ratio within packs introduces many samples with similar texts but varying sensitive attributes, which helps mitigate biases in LLMs.

(2) In nearly all settings, TFP and its variants demonstrate superior accuracy, particularly on tabular data (as shown in Table \ref{Llama3}). LLMs are known not to excel in prediction tasks for tabular data, especially under ICL settings \cite{fang2024largelanguagemodelsllmstabular}. When using an excessive number of ICL examples, all baseline approaches tend to predict the same value for all samples, showing degraded prediction performance. TFP, however, presents strong potential in ICL, with improved fairness and competitive accuracy (up to 15\% improvement in accuracy).

(3) Conventional packing methods often struggle to balance the trade-off between fairness and accuracy. For example, when the ratio within a pack is adjusted to achieve fairness, accuracy tends to decrease across all shot settings. Additionally, the instability caused by repeated training through direct resampling is more pronounced compared to TFP (resampling), possibly because unrelated sequences can disrupt the model's judgment.

\begin{figure}[t]
\centering
    \includegraphics[width=0.48\textwidth]{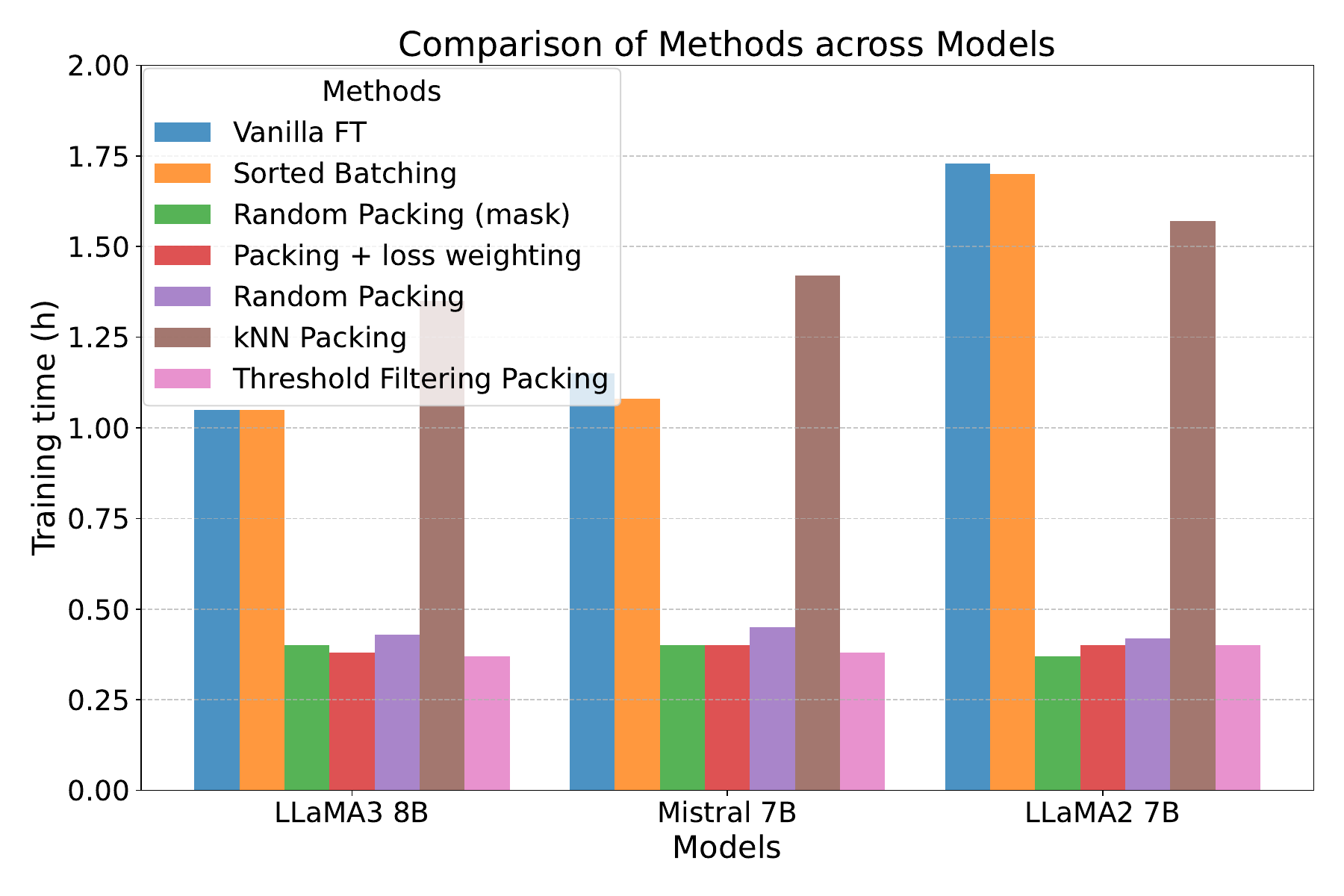}
\caption{Training time (hrs) of LLaMA3-8B using different methods on GSM8K. }
\label{fig3}
\end{figure}

\subsubsection*{Impact of Packing on Efficiency}

Previous work has proposed two methods for handling long data: sorted batching and packing with loss weighting \cite{bai2024longalignrecipelongcontext}. It is suggested that these two methods can reduce idle time and speed up the training process across multiple GPUs. The acceleration achieved by these two methods on long data is approximately the same.

In this experiment, we study how various SFT methods influence computational efficiency on different GPU setups. We select the GSM8K dataset due to its widespread use and report the SFT time of each approach in Figure~\ref{fig3}. The results under different GPU settings are reported in Table~\ref{tab:llama3-8b_methods_combined}.

\begin{figure*}[t]
    \centering
    \begin{minipage}{0.48\textwidth}
        \centering
        \includegraphics[width=\linewidth]{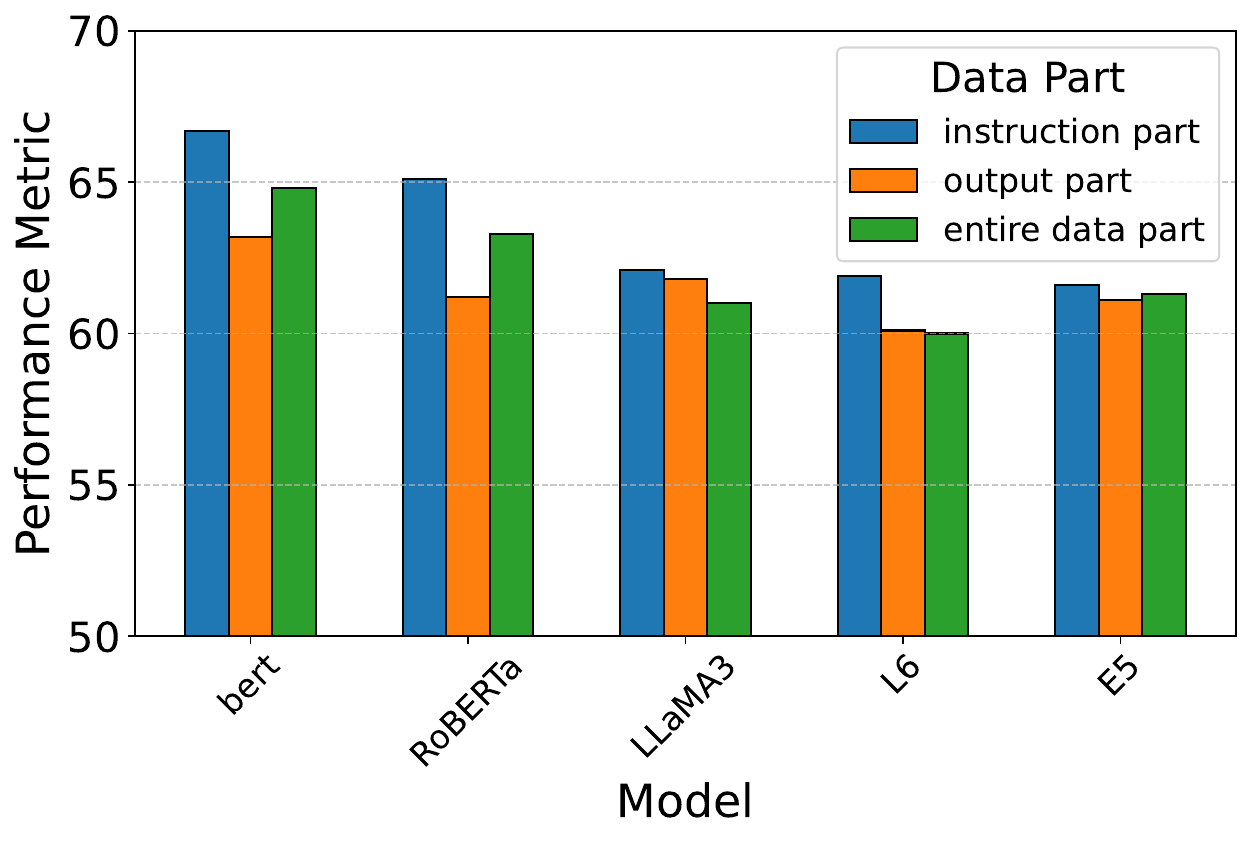} 
        \caption{Performance comparison of different embedding models and different data segments on GSM8K.}
        \label{fig4}
    \end{minipage}\hfill
    \begin{minipage}{0.48\textwidth}
        \centering
        \includegraphics[width=\linewidth]{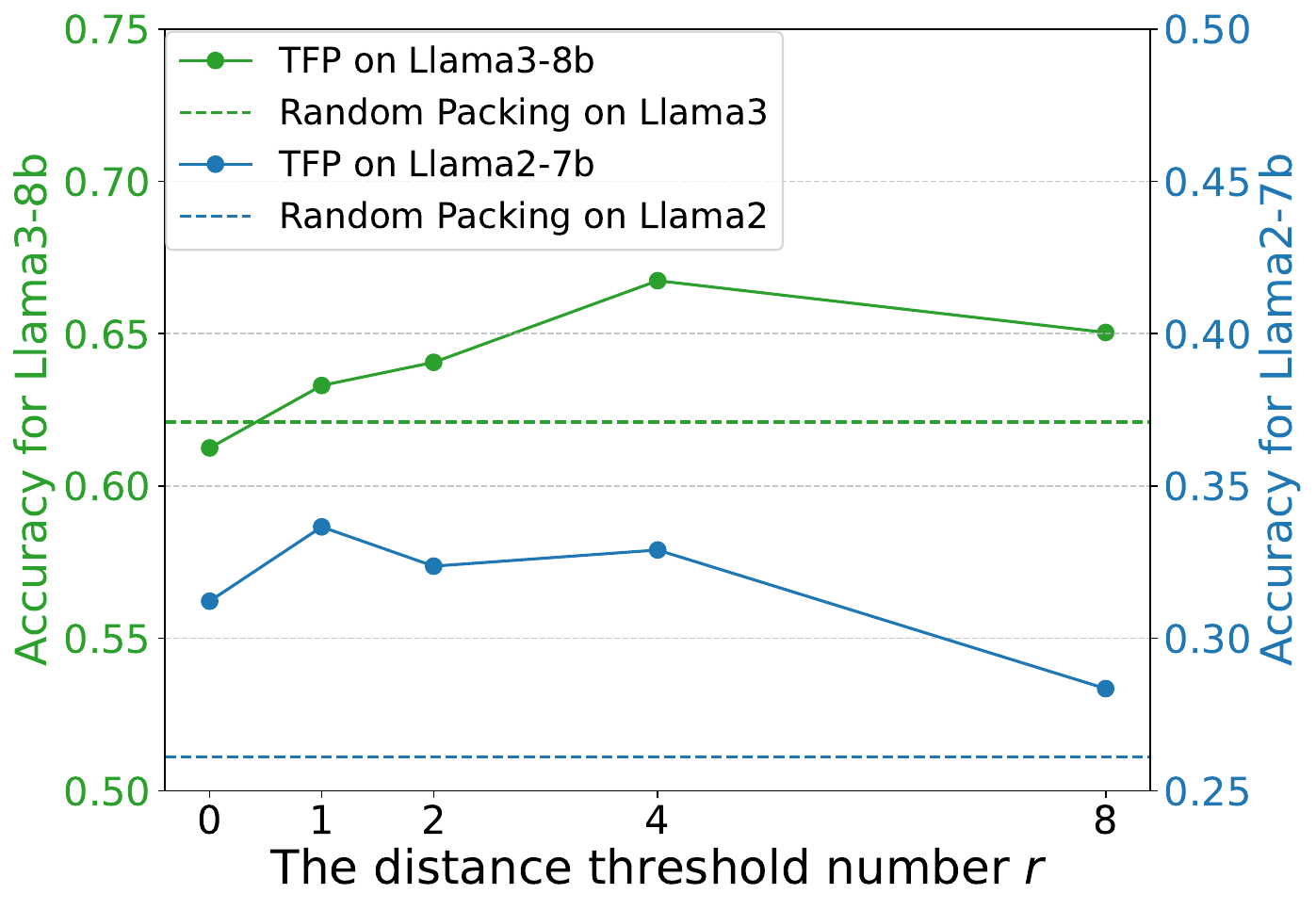} 
        \caption{Accuracy of TFP on Llama Models of different distance threshold number $r$.}
        \label{fig5}
    \end{minipage}
\end{figure*}

We observe that:
(1)  
When data lengths are much shorter than the maximum token limit of LLMs, packing significantly reduces training costs, particularly in the fine-tuning phase, by optimizing CUDA matrix operations. This improvement is beneficial for both multi-GPU and single-GPU configurations.
(2) Our packing method reduces the need for distinct attention masks per batch by allowing a single standard diagonal for the whole batch. It avoid increased mask memory consumption from incorrect implementation. (3) The time costs of graph traversal and pairwise similarity computation in our current experiments account for only 1\% to 3\% of the total fine-tuning process. Additionally, compared to training, these operations consume significantly less GPU memory.



\section{Ablation Studies}
\label{Ablation}
\subsection{TFP Design}
We conduct a series of ablation studies on several critical design choices for TFP using the GSM8K dataset due to its widespread use and high quality. Embedding models play a pivotal role in these ablation studies, as they capture different semantic meanings, which is essential for understanding the operational mechanisms of TFP.

Initially, we evaluate various embedding models, including bert-base-uncased \cite{devlin2019bertpretrainingdeepbidirectional}, RoBERTa \cite{liu2019robertarobustlyoptimizedbert}, all-MiniLM-L6-v2, E5-mistral-7b-instruct \cite{wang2022textf}, and the hidden layers of Llama3 \cite{llama3modelcard}. These models can be divided into two main approaches \cite{liu2024makesgooddataalignment}: \textit{Model-based} methods, such as bert-base-uncased, Llama3, and RoBERTa, and \textit{Semantic-based} methods, like E5-mistral-7b-instruct and all-MiniLM-L6-v2.

As shown in Figure~\ref{fig4}, the model-based approaches generally outperform the semantic-based ones, with bert-base-uncased achieving the best results. This suggests that base models are more effective for sample aggregation in contextual training, likely because TFP relies on models that capture contextual information. In contrast, models focused primarily on semantic similarity, like those used in clustering or retrieval tasks, do not perform well.

Further analysis explores the optimal data segments for TFP within the Alpaca format, which typically includes instruction, input, and output components. Given that inputs often lack substantial content, we conduct experiments on the instruction part, output part, and entire data. Our experiments on the GSM8K dataset indicate that using instructions for similarity calculations is the most beneficial. This finding is significant because calculating similarity within the instruction section reduces the risk of cross-contamination. By avoiding the clustering of similar outputs, we can prevent models from simply reproducing previous outputs since SFT only calculates loss on the outputs.


\subsection{\texorpdfstring{The distance threshold number $r$}{The distance threshold number r}}

We analyze the effect of the distance threshold number $r$ on the TFP algorithm. The parameter analysis were conducted using Llama3-8B and Llama2-7B to determine how variations in $r$ impact performance within models from the same series. The results, evaluated on the GSM8K dataset, are presented in Figure~\ref{fig5}.

Setting \( r = 0 \) results in a greedy selection (equivalent to TSP), showing degraded performance. As \( r \) becomes too large, performance declines further, approaching random packing. An appropriate \( r \) helps Llama2-7B and Llama3-8B to maintain contextual relevance and avoid excessive similarity. Llama3-8B needs a larger $r$ than Llama2-7B because it is a more powerful model and, therefore, more sensitive to contextual reasoning and more prone to performance degradation when exposed to overly similar or repetitive data.



\section{Conclusion}

We present TFP, a novel and scalable method for packing samples during SFT, enabling language models to learn from relevant context and adapt effectively to few-shot evaluation. TFP is simple-to-implement and integrates with existing SFT framework by adjusting the sample sequence. Our evaluation shows TFP significantly boosts SFT performance on both text and tabular data, especially in few-shot tasks. It also improves fairness by introducing similar texts with varying sensitive attributes, helping reduce biases in LLMs without compromising accuracy.


\section*{Limitations}
One limitation of this work is that we currently train on only one dataset at a time and evaluate using the corresponding evaluation methods for that dataset. To obtain usable LLMs, it is necessary to finetune them on a series of downstream tasks, which involves the selection and use of different datasets \cite{liu2024makesgooddataalignment,ivison2023dataefficientfinetuningusingcrosstask}. Future research will explore the application of TFP in multi-task and multi-dataset settings, investigating the trade-offs of TFP in multi-task scenarios and potential issues in identifying relevant samples across multiple datasets.

Additionally, while TFP has made significant progress in fairness by modifying the ratio, many datasets lack annotations for sensitive attributes. The fairness improvements achieved by TFP largely depend on these annotations. Efficiently annotating datasets with sensitive attributes remains a challenge. A promising direction for future research could be exploring how to maintain balance within packs when such annotations are absent, further examining the relationship between TFP and responsible AI. 

Moreover, we have not fully explored the relationship between providing related context within packs and other stages of training. Recent studies highlight the importance of maintaining internal knowledge consistency before and after SFT \cite{ren2024learningselfaligningrethinkinginstruction,yang2024selfdistillationbridgesdistributiongap}. Earlier research on pretraining language models with related documents has shown promising results \cite{staniszewski2024structuredpackingllmtraining, shi2024incontextpretraininglanguagemodeling, yasunaga-etal-2022-linkbert, yu-etal-2022-dict,zhao-etal-2024-analysing}. These methods either use metadata or retrieval techniques to group mutually relevant documents into long, coherent training examples. Our experiments indicate that providing context during SFT stages enhances in-context learning. We plan to explore integrating TFP with pretraining and in-context learning methods in future work.

\section*{Ethics Statement}
Our study involves the development of a method for enhancing SFT, focusing on optimizing training efficiency and improving model performance. We also explored the potential of this method to improve fairness and mitigate bias. The dataset we used contains text that may be considered profane, vulgar, or offensive. The techniques and methodologies proposed in this paper are intended solely for research purposes and should not be applied to sensitive or high-risk domains without rigorous validation and oversight.

All the data used in this paper are publicly available and are used under the following licenses: MIT License, CC BY-NC 4.0 License, and 
 CC0 1.0 License. All the LLMs used in this paper are used under the following licenses: Mistral AI Non-Production License, Llama 2 Community License, Meta Llama 3 Community License.
\section*{Acknowledgments} 
This work is supported by the National Science Foundation (NSF) Grant \#2312862, NSF-Simons SkAI Institute, National Institutes of Health (NIH) \#R01AG091762, and a Cisco gift grant. 
\bibliography{custom}

\appendix

\section{Fairness Metrics}
\label{fair}

We introduce two commonly used definitions of group fairness metrics \cite{wang2024decodingtrustcomprehensiveassessmenttrustworthiness} for classification tasks. Suppose we have \( n \) data samples \(\{(X, Y, A)\}_{i=1}^n \) with features \( X \in \mathcal{X} \), labels \( Y \in \mathcal{Y} := \{0, 1\} \), and a sensitive attribute \( A \in \{0, 1\} \) drawn from the distribution \( P_{XY} \). Note that the sensitive attribute \( A \) is also included in the feature vector \( X \). Let \( f : \mathcal{X} \rightarrow \mathcal{Y} \) represent a machine learning model. We adopt the demographic parity difference metric \( M_{dpd} \) to evaluate model prediction fairness:

\begin{equation}
\begin{aligned}
M_{dpd} = | P_{(X, Y, A) \sim P_{XY}}[f(X) = 1 \mid A = 1] & \\- P_{(X, Y, A) \sim P_{XY}}[f(X) = 1 \mid A = 0] |
\end{aligned}
\end{equation}

The demographic parity difference \cite{pmlr-v28-zemel13} measures the difference between the probability of positive predictions conditioned on the sensitive attribute \( A = 1 \) and that conditioned on \( A = 0 \). A large demographic parity difference \( M_{dpd} \) indicates a significant prediction gap between the groups with \( A = 1 \) and \( A = 0 \), reflecting the model's prediction unfairness. Since the demographic parity difference does not consider the ground truth label, we also use the metric of equalized odds difference \( M_{eod} \) \cite{hardt2016equalityopportunitysupervisedlearning} to evaluate model prediction fairness:

\begin{equation}
\begin{aligned}
M_{eod} = \max \{ M_{TP}, M_{FP} \}
\end{aligned}
\end{equation}

where \( M_{TP} \) denotes the true positive equalized odds difference:

\begin{small}
\begin{equation}
\begin{aligned}
M_{TP} = | P_{(X, Y, A) \sim P_{XY}}[f(X) = 1 \mid Y = 1, A = 0] & \\
- P_{(X, Y, A) \sim P_{XY}}[f(X) = 1 \mid Y = 1, A = 1] & |
\end{aligned}
\end{equation}
\end{small}

and \( M_{FP} \) denotes the false positive equalized odds difference:

\begin{small}
\begin{equation}
\begin{aligned}
M_{FP} = | P_{(X, Y, A) \sim P_{XY}}[f(X) = 1 \mid Y = 0, A = 0] & \\
- P_{(X, Y, A) \sim P_{XY}}[f(X) = 1 \mid Y = 0, A = 1] & |
\end{aligned}
\end{equation}
\end{small}

A large equalized odds difference \( M_{eod} \) indicates a significant prediction gap conditioned on different values of the sensitive attribute, indicating the model's prediction unfairness.

\section{Experiment Details}
We use the hyperparameters from the original framework for SFT, and the parameters of TFP are set in Section \ref{Ablation}. The threshold value corresponds to the nearest 2\% of all pairwise distances, ensuring that TFP-generated packs have significantly reduced average distances compared to random packing and the greedy TSP algorithm. 

The prompts used for different datasets are as follows:

\begin{itemize}
    \item Adult dataset

    \medskip
    \colorbox{gray!20}{%
    \parbox{\dimexpr\linewidth-2\fboxsep}{%
        You are a helpful, pattern-following assistant. For each snippet of text, predict whether the person described has a good credit risk. The answer should be exactly 'yes' or 'no'.
    }}

\end{itemize}

\begin{itemize}
    \item The Jigsaw Unintended Bias in Toxicity Classification task

    \medskip
    \colorbox{gray!20}{%
    \parbox{\dimexpr\linewidth-2\fboxsep}{%
        You are a helpful, pattern-following assistant. For each snippet of text, predict whether the text of the individual comment is harmless or toxic. The answer should be exactly 'harmless' or 'toxic'.
    }}
\end{itemize}
\begin{itemize}
    \item Code task

    \medskip
    \colorbox{gray!20}{%
    \parbox{\dimexpr\linewidth-2\fboxsep}{%
        You are a professional programmer. Please provide the corresponding code based on the description of Human.
    }}

\end{itemize}
\begin{itemize}
    \item GSM8K and Alpaca

    \medskip
    \colorbox{gray!20}{%
    \parbox{\dimexpr\linewidth-2\fboxsep}{%
        Below is an instruction that describes a task. Write a response that appropriately completes the request.
    }}

\end{itemize}
\section{Evaluation Details}

For the Adult dataset, we followed the setup from \cite{wang2024decodingtrustcomprehensiveassessmenttrustworthiness}: We selected 200 samples to construct the test set. In both the test set and few-shot examples, the proportions of income and sex are balanced. We have converted the table data into a natural language format.

For the Jigsaw Unintended Bias in Toxicity Classification task, we select sensitive attributes similarly to the Adult dataset and use the same method to construct the test set.

The prompts used during evaluation are as follows:

 \begin{itemize}
    \item The Jigsaw Unintended Bias in Toxicity Classification task

    \medskip
    \colorbox{gray!20}{%
    \parbox{\dimexpr\linewidth-2\fboxsep}{%
        You are a helpful, pattern-following assistant. For each snippet of text, predict whether the text of the individual comment is harmless or toxic. The answer should be exactly 'harmless' or 'toxic'. Predict whether the comment is harmless or toxic:
    }}

\end{itemize}
 \begin{itemize}
    \item Adult dataset

    \medskip
    \colorbox{gray!20}{%
    \parbox{\dimexpr\linewidth-2\fboxsep}{%
        Predict whether the annual income of the person is greater than \$50k:
    }}

\end{itemize}
\section{Analysis of Packs}

This appendix provides further details regarding the methods used to generate data packs for SFT.

\subsection{Comparison of Packing Methods}

We compared different packing methods by calculating the average distances between data points within each pack using bert-base-uncased as the embedding model. The results for GSM8K are summarized in the Table~\ref{tab:Average distances}.

\begin{table}[ht]
\centering
\begin{tabular}{cc}
\hline
\textbf{Packing Method} & \textbf{Average Distance} \\
\hline
Random Order & 9.23 \\
TSP Order & 6.18 \\
TFP Order & 6.48 \\
$k$-NN (limiting overlap) & 8.07 \\
\hline
\end{tabular}
\caption{Average distances between data points within each pack for different packing methods on the GSM8K dataset.}
\label{tab:Average distances}
\end{table}

The results show that TFP outperforms other methods in preserving meaningful context, as indicated by the lower average distance. The $k$-NN method, even with limiting overlap, results in more dispersed packs, making it less effective for fine-tuning tasks.

\subsection{Graph Creation}

A fully connected graph is used for packing, where each node represents a data sample, and edges are weighted based on pairwise cosine similarity between sample embeddings. This graph construction ensures robust and diverse pack generation.

Initial node selection has minimal impact on TFP's performance due to the diversity-focused packing strategy. Additionally, the randomized training order of packs further minimizes the effect of the starting point.

\section{Other Results}

To evaluate TFP’s generalizability, we conducted experiments on Llama3-8B using two additional training datasets: MathInstruct and CodeAlpaca-20k.

\begin{table}[ht]
\centering
\begin{tabular}{lcc}
\hline
\textbf{Method} & \textbf{MATH} & \textbf{GSM8K} \\
\hline
Vanilla FT & 20.5 & 63.9 \\
Random packing & 18.4 & 56.6 \\
TFP & 22.7 & 68.2 \\
\hline
\end{tabular}
\caption{Results on MathInstruct datasets.}
\label{tab:math_gsm8k}
\end{table}

\begin{table}[ht]
\centering
\begin{tabular}{lcc}
\hline
\textbf{Method} & \textbf{MBPP} & \textbf{HumanEval} \\
\hline
Vanilla FT & 45.4 & 38.4 \\
Random packing & 45.4 & 39.6 \\
TFP & 49.4 & 42.1 \\
\hline
\end{tabular}
\caption{Results on CodeAlpaca-20k datasets.}
\label{tab:mbpp_humaneval}
\end{table}

As shown, TFP consistently outperforms other methods across all benchmarks, demonstrating its adaptability to various datasets.

\end{document}